# Predicting Short-Term Energy Demand in the Smart Grid: A Deep Learning Approach for Integrating Renewable Energy Sources in Line with SDGs 7, 9, and 13

**Md Saef Ullah Miah[1], Junaida Sulaiman[2], Md. Imamul Islam[3], Md Masuduzzaman[4], Molla Shahadat Hossain Lipu[5], and Ramdhan Nugraha[6]**

**[1]Department of Computer Science, Faculty of Science and Technology, American International University-Bangladesh (AIUB), Dhaka, Bangladesh.**
**[2]Faculty of Computing, Universiti Malaysia Pahang Al-Sultan Abdullah, Pekan, Malaysia.**
**[3]FTKEE, Universiti Malaysia Pahang Al-Sultan Abdullah, Pekan, Malaysia.**
**[4]Department of IT Convergence Engineering, Kumoh National Institute of Technology.**
**[5]Department of Electrical and Electronic Engineering, Green University of Bangladesh, Dhaka, Bangladesh.**
**[6]Faculty of Electrical Engineering, Telkom University, Bandung 40257, Indonesia.**

Corresponding author:
Ramdhan Nugraha[1]

Email address: ramdhan@telkomuniversity.ac.id

## ABSTRACT

Integrating renewable energy sources into the power grid is becoming increasingly important as the world moves towards a more sustainable energy future in line with SDG 7. However, the intermittent nature of renewable energy sources can make it challenging to manage the power grid and ensure a stable supply of electricity, which is crucial for achieving SDG 9. In this paper, we propose a deep learning model for predicting energy demand in a smart power grid, which can improve the integration of renewable energy sources by providing accurate predictions of energy demand. Our approach aligns with SDG 13 on climate action, enabling more efficient management of renewable energy resources. We use long short-term memory networks, well-suited for time series data, to capture complex patterns and dependencies in energy demand data. The proposed approach is evaluated using four historical short-term energy demand data datasets from different energy distribution companies, including American Electric Power, Commonwealth Edison, Dayton Power and Light, and Pennsylvania-New Jersey-Maryland Interconnection. The proposed model is compared with three other state-of-the-art forecasting algorithms: Facebook Prophet, Support Vector Regression, and Random Forest Regression. The experimental results show that the proposed REDf model can accurately predict energy demand with a mean absolute error of 1.4%, indicating its potential to enhance the stability and efficiency of the power grid and contribute to achieving SDGs 7, 9, and 13. The proposed model also has the potential to manage the integration of renewable energy sources effectively.

## INTRODUCTION

The world is currently experiencing a serious dilemma in the energy field due to the rapid depletion of fossil fuels due to rising populations, urbanization, and technological advancements. In addition, burning fossil fuels results in water and air contamination, climate change, and the production of greenhouse gases, all of which contribute to the acceleration of global warming and have severe adverse effects on ecosystems and human health (28). To mitigate the impacts of climate change, scientists, academics, and policymakers are working to mainstream renewable energy (RE) as a replacement for carbon-based power sources (27). To achieve the 1.5°C scenario goal by 2050, the most significant threshold is to ensure 90% electricity generation from RE sources and 79% of the overall energy consumption (26). Over the last decade, the installation and generation of renewable energy in terms of off-grid and on-grid systems have increased significantly. According to the International Renewable Energy Agency (IRENA), the latest trends in renewable energy are shown in Fig. 1 (3). This figure shows that the maximum number of installations get direct connections to the grid, and the increase in solar and wind-based power plants is noticeable compared to other technologies.

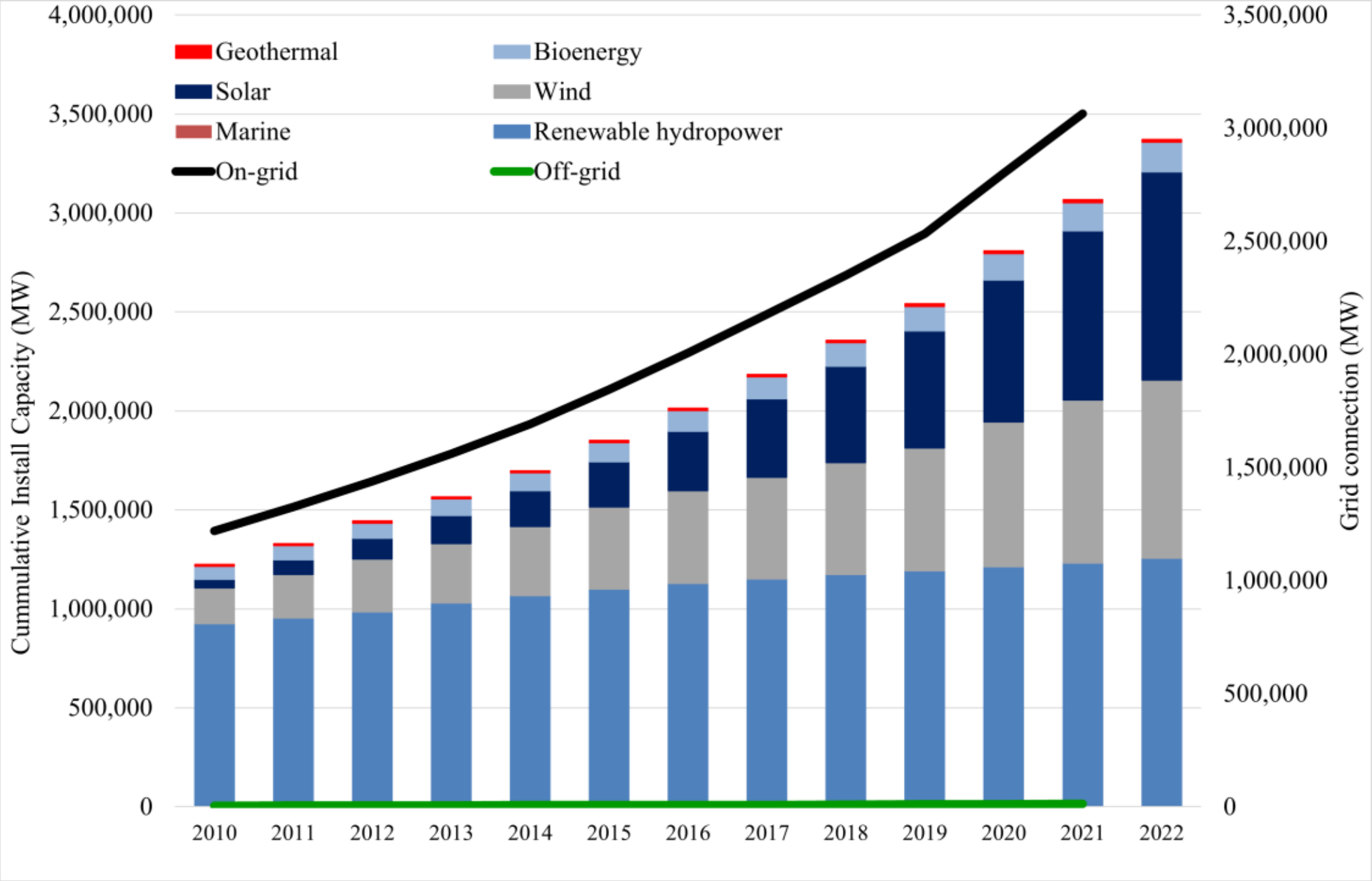

**Figure 1.** Global renewable energy trend and grid application scenario

## Background

The imperative transition towards a sustainable future necessitates integrating renewable energy sources into power grids, reshaping global energy systems (5; 31). Aligned with the UN's ambitious goal of universal access to affordable, reliable, and modern energy by 2030 (Martin; 22), this integration emerges as pivotal for sustainable development. The UN Sustainable Development Goals (SDGs) set the framework, with SDG 7 (Affordable and Clean Energy), SDG 9 (Industry, Innovation and Infrastructure), and SDG 13 (Climate Action) standing as critical benchmarks for an equitable and environmentally sustainable energy landscape (11).

Deep learning models play a crucial role in achieving these SDGs, particularly in predicting energy demand with precision. This accuracy enables optimal management of renewable resources, diminishes reliance on fossil fuels, reduces greenhouse gas emissions, and fortifies resilient and efficient infrastructure. Achieving high shares of renewable energy, as emphasized by SDG 7, diversifies the energy mix, decreases costs, and enhances accessibility.

Extending the focus to SDG 9, the study underscores the importance of fostering sustainable infrastructure and promoting innovation in the energy sector. Additionally, SDG 13 becomes a compelling imperative as the transition to renewable energy significantly contributes to mitigating the impacts of climate change.

Despite the promise of renewable sources, challenges persist, notably the unpredictability of energy generation (6; 36). Addressing this challenge requires accurate predictions of energy demand, aligning with SDG 7. Developing effective predictive models becomes paramount in realizing these objectives.

This study delves into the intricate landscape of facilitating greater integration of renewable energy by predicting aggregated energy demand, aligning with multiple SDGs crucial for constructing an equitable and sustainable energy future (10).

Harnessing clean energy sources such as hydroelectricity, geothermal, biomass, solar, and wind power not only reduces energy prices but also assures a dependable and sustainable energy supply globally (42; 15). Addressing climate change through the grid integration of renewable energy becomes pivotal for reducing greenhouse gas emissions (19) and mitigating its adverse effects (32). The transition from fossil fuels to renewable energy stands as a tangible step in fulfilling climate commitments under the Paris Agreement, safeguarding the planet for future generations (4).

While the integration of renewable sources holds promise, challenges persist, notably the unpredictability of energy generation (6; 36). Tackling this challenge necessitates predicting energy demand using deep learning for a smart power grid, reflecting SDG 7. The development of accurate predictive models is a critical enabler in achieving this.

## Literature Review

Machine learning techniques have garnered considerable attention in the energy sector recently. Within this evolving landscape, several noteworthy innovations have come to the forefront. For instance, an innovative hybrid deep learning model, which combines Long Short-Term Memory (LSTM) (53) neural networks with the stationary wavelet transform, has demonstrated its ability to enhance predictions of household energy consumption. This model addresses challenges associated with irregular behaviors and univariate data issues (52). Additionally, in the context of developing countries, the application of artificial neural networks (ANNs) (20) coupled with meta-heuristic techniques has proven effective for forecasting load demand. This approach is pivotal in guiding energy-efficient growth in regions where technical resources and infrastructure are limited (9). Furthermore, a novel framework has been introduced to streamline residential energy management. This framework integrates an ANN-based forecast engine with a controller using the Differential Evolution Algorithm modified for Grey Wolves (DA-GmEDE). The outcome is a substantial reduction in energy costs and peak-to-average ratios, emphasizing the potential for cost-effective and efficient energy management within the smart grid (21). These groundbreaking innovations underscore the pivotal role that machine learning plays in shaping a sustainable and efficient energy landscape.

The authors of the paper (8) analyzed the efficiency of different deep learning models, including RNN, LSTM, and GRU, in predicting energy demand for the Smart Grid Smart City project using datasets from 2010 to 2014. The models are evaluated using RMSE, MAE, and $R^2$ scores. The results showed that GRU outperforms basic RNN and LSTM with the lowest RMSE error and highest $R^2$ score due to its ability to deal with the vanishing gradient problem and its impact on the number of parameters.

In another work (7), the authors proposed using hybrid deep learning methods to improve load forecasting accuracy in the Saudi smart grid system. It aims to develop reliable forecasting models and understand the relationship between various features and attributes in the Saudi smart grid. The model uses a real dataset from Jeddah and Medinah grids for an entire year with a one-hour time resolution and compares prediction results with conventional deep learning methods, including RNN (24), LSTM (23), GRU (16), and CNN (34). The results show that the proposed hybrid deep learning models, particularly CNN-GRU and CNN-RNN, provide 1.4673% and 1.222% improvement in load forecasting accuracy, respectively, compared to the benchmark strategy.

In the work (47), the authors used LSTM-RNN-based deep learning architecture to predict household electrical energy consumption two months in advance. The model is trained on relevant features and evaluated by comparing actual and predicted values. The proposed model helps households conserve energy and is evaluated using the UCI repository dataset of domestic electric consumption (17). The results showed that the LSTM model has much higher precision than statistical and engineering prediction models, with a compatible RMSE of 0.6 compared to conventional models.

The study by Taleb et al. (49) proposed a hybrid machine learning model that combines standard neural networks with an automatic weight update process based on past errors. This flexible model can predict energy demand over various time ranges and regions. The effectiveness of the proposed model was demonstrated by achieving a mean absolute error (MAE) of 372.08 in energy demand prediction for Mayotte Island.

Residential load forecasting is becoming increasingly important as smart meters are increasingly deployed at the household level to collect historical data on energy consumption. In the study by Mubashar et al. (38), they proposed a method for load forecasting and validated it using real-world data sets. They compared the performance of their proposed method, which uses LSTM models, with two commonly used techniques, ARIMA and exponential smoothing. They evaluated the accuracy of load forecasts generated using these three techniques using real data from 12 houses over three months. The results indicated that LSTM models performed better than the other two methods for time series-based predictions. Their model achieved an MAE of 2.44736176.

Another work by Rosato et al. (46) presented a novel deep learning approach for multivariate prediction of energy time series. The proposed approach utilized Convolutional Neural Network and Long Short-Term Memory models to combine and filter several correlated time series while considering their long-term dependencies. The learning scheme is implemented as a stacked deep neural network, with one or more layers feeding their output into the input of the subsequent layer. The effectiveness and accuracy of the proposed approach are demonstrated through real-world applications in the energy sector, highlighting its robustness and accuracy. The lowest RMSE the method achieved among all the variations tested is 2.252, achieved on a baseline 1-day forecast.

Electric power load demand forecasting is critical for energy management, requiring accurate planning and infrastructure investment predictions. Despite a lot of research in this area, accuracy remains an issue. The study by Nguyen et al. (39) proposed an electricity demand forecasting method based on the LSTM deep learning model, tested using six years of power consumption data in Vietnam. The proposed method achieved an RMSE of 9.63, indicating its potential as a valuable tool for energy sector studies.

This paper by Pramono et.al (44), proposed a method for short-term load forecasting using a wavenet-based model that employs dilated causal residual convolutional neural network (CNN) and long short-term memory (LSTM) layers. The proposed model outperforms other deep learning-based models in terms of root mean squared error (RMSE), and mean absolute error (MAE), achieving RMSE and MAE equal to 203.23, and 142.23 for the ENTSO-E testing dataset 1, and 292.07 and 196.95 for ENTSO-E dataset 2. For the ISO-NE dataset, the RMSE, and MAE are equal to 85.12, 58.96 for ISO-NE testing dataset 1 and 85.31, and 62.23 for ISO-NE dataset 2. The proposed method aimed to support the demand response program in hybrid energy systems, especially those using renewable and fossil sources. Two different ways of conducting model testing were conducted: one using datasets with identical distributions as the validation data and the other with unknown distributions.

This paper by Li et al. (35) proposes a novel non-intrusive attention-augmented deep learning model called NAP-BiLSTM for short-mid term electricity consumption prediction. The model comprises a non-intrusive attention-augmented-based preprocessing (NAP) block and a regular bidirectional long short-term memory (BiLSTM). The NAP block is a plug-and-play attention mechanism for time series that can be combined with a deep neural model without modifying its structure. The effectiveness of the proposed model is verified by two experiments, including univariate prediction using EC data from the U.S. and multivariate prediction using weather and energy data from Valencia. The results demonstrate that the proposed model outperforms state-of-the-art deep learning models regarding root mean square error, mean absolute error, and mean absolute percentage error for univariate and multivariate analyses. The paper reports that the proposed NAP-BiLSTM model achieved an improvement in MAE that varies from 2.36% to 16.52% compared to four baseline models (AM-BiLSTM, Stacked-BiLSTM, Adaboost, and attention-CNN-LSTM) for mid-term multidimensional time series prediction.

The performance of different models studied in different works is summarized in Table 1. This table shows the comparison of the studies with respect to different evaluation metrics. From the table, it can be observed that the LSTM-RNN (47) model did not report the MAE and $R^2$ values mentioning only RMSE. The CNN-GRU (7) model achieved a high $R^2$ score but did not report the MAE and RMSE values. On the other hand, the LSTM and GRU models reported MAE and $R^2$ values, but their performance was inferior to the proposed model. Apart from these, most studies did not report $R^2$ values.

The related works in this field demonstrate various approaches for predicting energy demand using machine learning techniques such as neural networks and time series analysis. These studies have shown promising results in improving energy demand forecasts, highlighting the potential of machine learning in the energy sector. However, there is still room for improvement, and further research is needed to refine and optimize these techniques to provide more accurate and reliable predictions.

## Problem Statement

Machine learning techniques have gained prominence in the energy sector, showcasing innovations in predicting energy demand. Noteworthy models, such as a hybrid approach combining LSTM networks with stationary wavelet transform, address irregular behaviors but leave gaps in providing a standardized evaluation framework (52). Similarly, the effectiveness of artificial neural networks (ANNs) in developing countries lacks comprehensive evaluation metrics, leaving uncertainties in their general applicability for load demand forecasting (9; 20).

**Table 1.** Performance comparison of different studies

| Reference | Year | Model | MAE | $R^2$ | RMSE |
|---|---|---|---|---|---|
| (35) | 2022 | NAP-BiLSTM | 0.02 | - | 0.027 |
| (8) | 2022 | LSTM | 0.021 | 0.53 | 0.039 |
| (8) | 2022 | GRU | 0.022 | 0.64 | 0.034 |
| (7) | 2022 | CNN-GRU | - | 0.973 | 0.816 |
| (47) | 2022 | LSTM-RNN | - | - | 0.6 |
| (49) | 2022 | CNN-LSTM-MLP | 372.08 | - | - |
| (38) | 2022 | LSTM | 2.447 | - | - |
| (46) | 2019 | CNN-LSTM | - | - | 2.252 |
| (39) | 2020 | LSTM | - | - | 9.63 |
| (44) | 2019 | CNN-LSTM-ENTSO-E | 142.23 | - | 203.23 |
| (44) | 2019 | CNN-LSTM-ISO-NE | 58.96 | - | 85.12 |

Novel frameworks integrating ANN-based forecast engines and controllers exhibit potential for cost-effective energy management, but challenges persist in refining forecasting models and understanding feature relationships within smart grids (21). The comparative analysis by Amalou et al. (8) reveals the superiority of GRU over RNN and LSTM, yet achieving accuracy in load forecasts remains a concern. The proposed hybrid deep learning models in the Saudi smart grid system show promise, but the need for further refinement to enhance forecasting accuracy is evident (7).

Studies predicting household energy consumption using LSTM-RNN architectures indicate higher precision but lack comprehensive reporting on metrics such as MAE and $R^2$ values, posing challenges in evaluating overall model performance (47). Taleb et al.'s (49) flexible energy demand prediction model shows effectiveness but raises questions about its adaptability across various contexts.

Residential load forecasting studies highlight the superior performance of LSTM models compared to traditional methods, yet discrepancies in reported metrics raise concerns about the robustness of these findings (38). Additionally, the multivariate prediction approach using CNN-LSTM models for energy time series lacks standardized benchmarks, hindering a comprehensive understanding of its effectiveness (46).

Despite advancements, the accuracy and reliability of energy demand predictions remain inconsistent across various models, revealing research gaps in standardization of evaluation metrics and the need for further model refinement. The absence of a unified framework hampers comparability and generalizability of results, necessitating comprehensive research to address these gaps and enhance the overall efficiency and sustainability of energy systems.

## Key Contributions

In alignment with recent studies and the defined objectives, this research introduces a novel approach utilizing deep learning, a sub-field of machine learning tailored for analyzing sequential data, particularly time series data (30; 25). The proposed methodology leverages a long short-term memory (LSTM) network, a type of recurrent neural network (RNN) explicitly designed to model sequential data with temporal dependencies (23). Training the model involves utilizing historical energy demand data, and its performance is rigorously evaluated using metrics such as mean absolute error (MAE), root mean squared error (RMSE), and coefficient of determination ($R^2$).

In addition to its prowess in accurately predicting energy demand, our proposed method showcases robust generalization capabilities for previously unobserved data. The ability to forecast energy demand accurately can revolutionize power grid management, fostering more efficient distribution and utilization of renewable energy sources while diminishing reliance on nonrenewable alternatives. Moreover, the proposed method has the capacity to enhance overall power infrastructure efficiency, cut costs, and facilitate the seamless integration of renewable energy sources.

The key contributions of this work extend beyond technical advancements to encompass a positive impact on Sustainable Development Goals (SDGs), particularly SDG 7 (Affordable and Clean Energy),

SDG 9 (Industry, Innovation, and Infrastructure), and SDG 13 (Climate Action). These contributions align with the United Nations' broader agenda for sustainable development. The specific contributions are as follows:

- **Introduction of LSTM-Based Deep Learning Model:** The proposed deep learning model addresses the pivotal SDG 7, ensuring affordable and clean energy for all. By enhancing the precision of energy demand forecasts, the model promotes efficient utilization of renewable energy sources, thus contributing to a sustainable and accessible energy future.

- **Minimization of Mean Absolute Error (MAE):** The emphasis on minimizing MAE in the energy demand prediction model directly supports SDG 9 by fostering innovation and infrastructure development. Accurate energy forecasts enable infrastructure optimisation, paving the way for more sustainable and resilient energy systems.

- **Comprehensive Model Evaluation:** The rigorous evaluation of the proposed approach using various metrics, including MAE, RMSE, and $R^2$, aligns with SDG 13. By improving the accuracy of energy demand predictions, the model contributes to climate action efforts, reducing reliance on nonrenewable energy sources and mitigating environmental impacts.

In essence, this paper's primary contributions lie in technical advancements and in addressing global challenges outlined in the SDGs. The research aligns with the United Nations' vision for a sustainable future, emphasizing the role of accurate energy demand forecasting in achieving key sustainable development objectives.

The remaining sections of this paper are structured as follows. The study's materials and methods, including data acquisition and pre-processing, model development, evaluation, and deployment, are described in detail in the following section. The experimental findings are then presented in the results and discussion section, along with a comparison to other existing methods. Finally, the study concludes with recommendations for future research.

## MATERIALS AND METHODS

In this study, we propose an approach based on deep learning for forecasting energy demand in a smart grid. The primary objective of this strategy is to improve the integration of renewable energy sources by providing accurate energy demand forecasts that can aid in the administration of the power grid. The proposed method continues with model evaluation and deployment, beginning with data collection, pre-processing, and model development. Starting with the formulation of the problem, we will describe each phase in detail and explain the tools and techniques used at each stage. The overview of the methodology employed in this study is shown in Fig. 2.

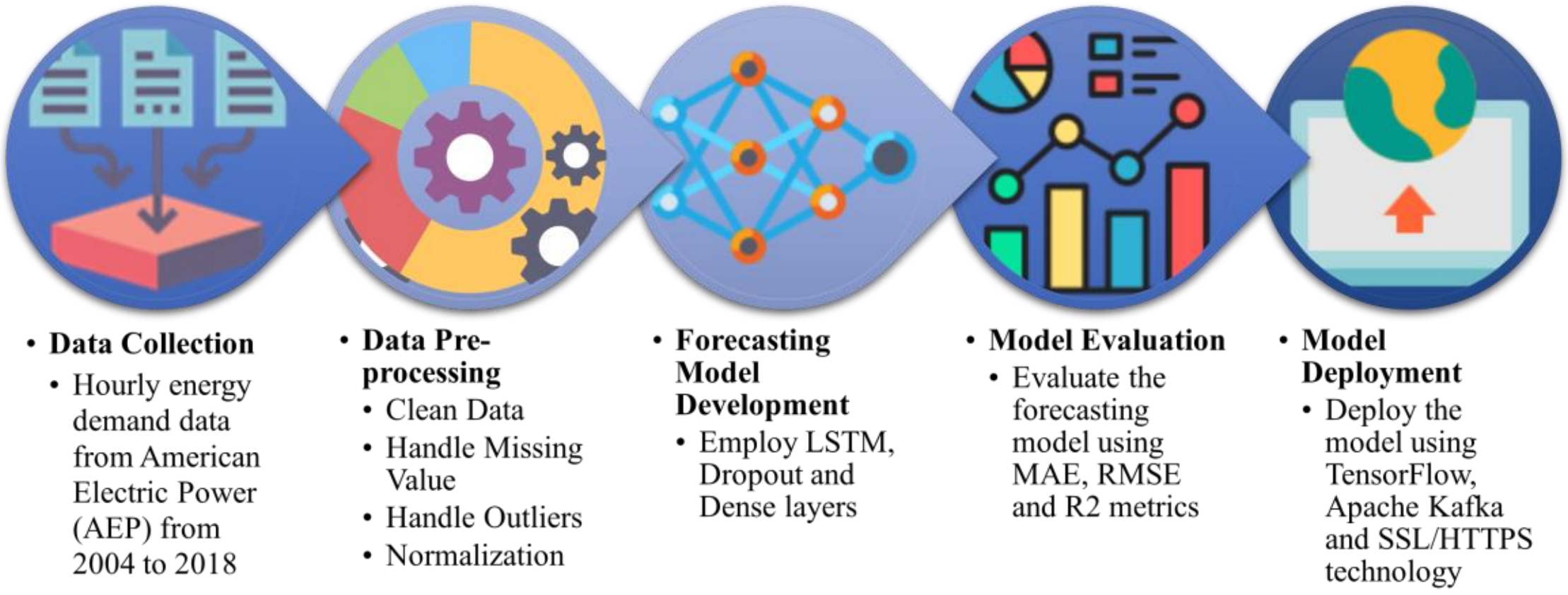

**Figure 2.** Methodology employed in this study

## Problem Formulation

The problem formulation section provides the foundation and context for the proposed solution in the study by clearly defining the objective, scope, and challenges of the energy demand forecasting problem. The problem of predicting energy demand for a smart power grid can be formulated mathematically as follows:

Given a time series of historical energy demand data, represented by a sequence of vectors $< D = d_1, d_2, \ldots, d_n >$, where $d_i$ is a vector of energy demand values at time $i$, the goal is to predict the energy demand at a future time, represented by a vector $d_{n+h}$, where $h$ is the number of time steps ahead for which the prediction is made.

This problem can be represented as a function $f(D) = d_{n+h}$, where the function $f$ maps the historical energy demand data to the predicted energy demand.

The objective is to find the optimal function $f$ that minimizes the prediction error, which can be defined as the mean absolute error (MAE) between the predicted and actual energy demand values.

## Data Collection

Data collection is essential in predicting energy demand in a smart power grid. This section briefly describes the steps and techniques used for data collection in the proposed study.

The initial phase of data collection involves the identification of pertinent data sources, necessitating the recognition of data types essential for enabling precise predictive modeling, including variables such as energy consumption, meteorological conditions, and economic indicators. Data acquisition encompasses a diverse array of potential sources, ranging from utility providers and government agencies to publicly accessible datasets. In the context of this research, the study leverages hourly energy demand data procured from American Electric Power (AEP), a prominent electric utility enterprise within the United States, renowned for its extensive service coverage encompassing over 5 million customers across 11 states (2). The dataset in question encompasses a robust repository comprising 121,273 data entries, documenting hourly energy consumption spanning from December 2004 to January 2018. Furthermore, to assess the model's performance comprehensively, supplementary datasets from COMED (13), DAYTON (40), and PJME (43) have been enlisted for evaluation. These datasets collectively serve as established benchmarks for scrutinizing the efficacy of energy demand forecasting models and are readily accessible through the following GitHub repository: https://github.com/panambY/Hourly_Energy_Consumption (41).

COMED hourly energy consumption data refers to the hourly electricity consumption data for the Commonwealth Edison (COMED) service area, which covers the northern part of Illinois in the United States. The data provides information on the hourly electricity demand for residential, commercial, and industrial customers in the COMED service area. This dataset contains historical energy consumption data in an hourly fashion from December 2011 to January 2018, with a total of 66497 data points.

DAYTON hourly energy consumption data refers to the hourly electricity consumption data for the Dayton Power and Light (DP&L) service area, which covers the city of Dayton, Ohio, and surrounding areas in the United States. The data provides information on the hourly electricity demand for residential, commercial, and industrial customers in the DP&L service area. This dataset contains historical energy consumption data in an hourly fashion from December 2004 to January 2018, with a total of 121275 data points.

PJME hourly energy consumption data refers to the hourly electricity consumption data for the Pennsylvania-New Jersey-Maryland Interconnection (PJM) regional transmission organization in the United States. The data provides information on the hourly electricity demand for a large portion of the eastern United States, covering 13 states and the District of Columbia, and includes variables such as date, time, temperature, and electricity demand. This dataset contains historical energy consumption data in an hourly fashion from December 2002 to January 2018, with a total of 145366 data points. Table 2 shows an overview of the datasets utilized to benchmark the proposed model.

After data has been collected, it has to be pre-processed to ensure that it is in a format compatible with the model. This includes removing any inconsistencies or errors from the data and transforming the data into a format that the model can use. The data pre-processing procedures described in the following section describe normalisation, feature scaling, and outlier removal.

## Data Pre-processing

Data pre-processing is essential in predicting energy demand in a smart power grid. It ensures that the data used to train and evaluate the model is clean, consistent, and in a suitable format for the model. This section will describe the steps and techniques used for data pre-processing in the proposed study.

**Table 2.** Overview of the datasets utilized in this study

| Dataset | Start Datetime | End Datetime | Data points |
|---|---|---|---|
| AEP | 2004-12-31 01:00:00 | 2018-01-02 00:00:00 | 121273 |
| COMED | 2011-12-31 01:00:00 | 2018-01-02 00:00:00 | 66497 |
| DAYTON | 2004-12-31 01:00:00 | 2018-01-02 00:00:00 | 121275 |
| PJME | 2002-12-31 01:00:00 | 2018-01-02 00:00:00 | 145366 |

The initial phase in data pre-processing is data cleansing. This includes eliminating any data inconsistencies, errors, or missing values. Common data cleansing techniques include removing duplicates, replacing absent values with imputed values, and converting data to a standard format.

The following phase transforms the data into a format the model can utilize. This includes the normalization, scaling, and encoding of categorical variables. Normalization is adjusting the data so that the mean is 0 and the standard deviations are 1. Scaling the data can prevent the magnitude of the data from affecting the model. The entire process of data pre-processing is presented in Algorithm 1.

**Algorithm 1** Data Pre-processing

1: **Input:** Raw Data $D_{raw}$
2: **Output:** Pre-processed Data $D_{pre}$
3: **Procedure:**
4: Load the raw data into the program: $D_{raw} \longleftarrow$load_data()
5: Check for missing values and handle them accordingly: $D_{pre} \longleftarrow$handle_missing_values($D_{raw}$)
6: Check for outliers and handle them accordingly: $D_{pre} \longleftarrow$ handle_outliers($D_{pre}$)
7: Normalize the data: $D_{pre} \longleftarrow$ normalize($D_{pre}$)
8: Divide the data into training and testing sets: ($D_{train}$, $D_{test}$) $\longleftarrow$split_data($D_{pre}$)
9: Return the pre-processed data: **return** $D_{pre}$

The algorithm serves as a systematic framework for preparing raw data, $D_{raw}$, for subsequent analysis. It takes the raw data as input and generates pre-processed data, $D_{pre}$, as output. The procedure unfolds as follows: First, it loads the raw data into the program using the function *load_data*(). Next, it checks for missing values within the dataset and employs appropriate handling procedures, yielding the pre-processed data, $D_{pre}$. Subsequently, the algorithm identifies and addresses outliers within $D_{pre}$ through the function *handle_outliers*(). Following this, the data is normalized using the *normalize*() function to ensure consistency and comparability among different features. The algorithm then partitions the pre-processed data into separate training and testing sets, denoted as ($D_{train}$, $D_{test}$), using the *split_data*() function. Finally, it returns the pre-processed data, $D_{pre}$, which is conditioned for further analysis.

Data pre-processing is a critical step in predicting energy demand in a smart power grid. It includes cleaning the data and transforming the data. These steps help ensure that the data used to train and evaluate the model is clean, consistent, and in a suitable format for the model.

## Model Development

This section describes developing a forecasting model to predict energy demand in a smart power grid. The proposed method employs a deep learning-based model, particularly a long short-term memory (LSTM) network, a recurrent neural network (RNN) designed to manage sequential data with temporal dependencies. LSTM is a form of recurrent neural network (RNN) architecture frequently used in deep learning applications for sequence modelling, including natural language processing, speech recognition, and time series forecasting.

In time series forecasting, LSTM is useful for capturing difficult-to-model long-term dependencies in the data. Predicting hourly energy demand is one possible application of LSTM in time series forecasting. The model is trained on historical data to use LSTM to predict hourly energy demand to understand the patterns and relationships between the input features and the target variable, in this case, energy demand in megawatts. Based on the input features, the model can predict future time steps.

LSTM comprises a set of nonlinear transformations that operate on the input and hidden states of the network, as well as gating mechanisms that regulate the passage of information through the network. The equations for a single LSTM cell are shown in equation (1)(48).

$$
\begin{aligned}
i_t &= \sigma(W_{xi}x_t + W_{hi}h_{t-1} + b_i) \\
f_t &= \sigma(W_{xf}x_t + W_{hf}h_{t-1} + b_f) \\
\tilde{C}t &= \tanh(Wxcx_t + W_{hc}h_{t-1} + b_c) \\
C_t &= f_t \odot C_{t-1} + i_t \odot \tilde{C}t \\
o_t &= \sigma(Wxox_t + W_{ho}h_{t-1} + b_o) \\
h_t &= o_t \odot \tanh(C_t)
\end{aligned}
\tag{1}
$$

Here, $x_t$ is the input at time step $t$, $h_{t_1}$ is the hidden state of the previous time step, $W$ and $b$ are the weights and biases of the network, and $\sigma$ and tanh are the sigmoid and hyperbolic tangent activation functions, respectively. The equations involve several gates, including an input gate $i_t$, a forget gate $f_t$, and an output gate $o_t$, which control the flow of information through the network. The cell state $C_t$ is updated based on the input and hidden states, and the hidden state $h_t$ is computed as a function of the cell state and the output gate.

The initial phase in model development is to collect and pre-process the data. As inputs to the model, we utilized historical energy demand data. The data was collected from the PJM data interface for the AEP zone and pre-processed to ensure it was in the correct format for the model. The pre-processing stages included data cleansing, handling missing values, and data normalization.

Next, the network structure of the LSTM model was defined. Input, multiple LSTM, dropout, and output layers comprise the LSTM model. The number of neurons in the input layer corresponds to the number of input features, which in this instance are the historical energy demand data. The LSTM layers are the basis of the model and are responsible for learning the temporal dependencies in the data. Multiple LSTM layers were layered on top of one another to improve the model's ability to learn complex data patterns. A grid search cross-validation technique was used to ascertain the number of LSTM layers and the number of neurons in each layer. Grid search cross-validation entails creating a grid of possible values for each hyperparameter and evaluating the model's performance on each combination of hyperparameters using a cross-validation strategy. Then, the model's performance is compared across all possible combinations, and the hyperparameters that yield the highest performance are chosen. This computationally intensive technique offers a systematic and trustworthy method for choosing optimal hyperparameters for deep learning models. Algorithm 2 presents the grid search method employed in this study. A single neuron in the output layer represents the anticipated energy demand.

**Algorithm 2** Grid Search for LSTM Hyperparameters

```
Input: Training data
Output: Best hyperparameters
Define the grid of hyperparameter values
Initialize variables for best hyperparameters and best performance metric
for each combination of hyperparameters do
    Build the LSTM model with the current hyperparameters
    Train and evaluate the model using cross-validation
    if model performance is better then
        Update the best hyperparameters and performance metric
    end if
end for
Retrieve the best hyperparameters
```

The LSTM model was implemented using the Keras (12) library in Python (51). The model was trained using an Adam optimizer and mean squared error (MSE) as the loss function. The model was trained for a fixed number of epochs, and the training process was stopped when the model's performance on a validation set stopped improving. Algorithm 3 presents the steps utilized in the forecasting model's development. The architecture of the proposed model is shown in Fig. 3.

**Algorithm 3** Proposed forecasting model

Initialize the model: model = Sequential()
Add an LSTM layer with x units: model.add(LSTM(x, input_shape=(timesteps, features)))
Add a dropout layer with rate of 0.1: model.add(Dropout(0.1))
Add an LSTM layer with x units: model.add(LSTM(x,Return_sequence='False'))
Add a dropout layer with rate of 0.1: model.add(Dropout(0.1))
Add a fully connected layer with y units: model.add(Dense(y))
Compile the model: model.compile(loss='mse', optimizer='adam')
Fit the model on the training data: model.fit(X_train, y_train, epochs=z, batch_size=w)
Make predictions on the test data: y_pred = model.predict(X_test)

Here, timesteps are the number of time steps in the input data, features are the number of features in the input data, x is the number of units in the LSTM layer, y is the number of units in the fully connected layer, z is the number of epochs for training, and w is the batch size for training. X_train and y_train represent the training data, and X_test represents the test data. In the proposed model, the unit used in LSTM is 200, and the number of units used in the fully connected layer is 1. The training epoch of the model is 10, and batch_size is 1000.

The algorithm starts by initializing the model using the Sequential() function from the Keras library, and the LSTM, dropout, and fully connected layers are added using the add() function. The model is then compiled using the compile() function and trained using the fit() function. Finally, predictions are made on the test data using the predict() function.

The trained model was then evaluated using performance metrics such as MAE, RMSE, and coefficient of determination ($R^2$) on a test set. These metrics were used to evaluate the model's ability to predict energy demand accurately. The model was then deployed and used to predict energy demand.

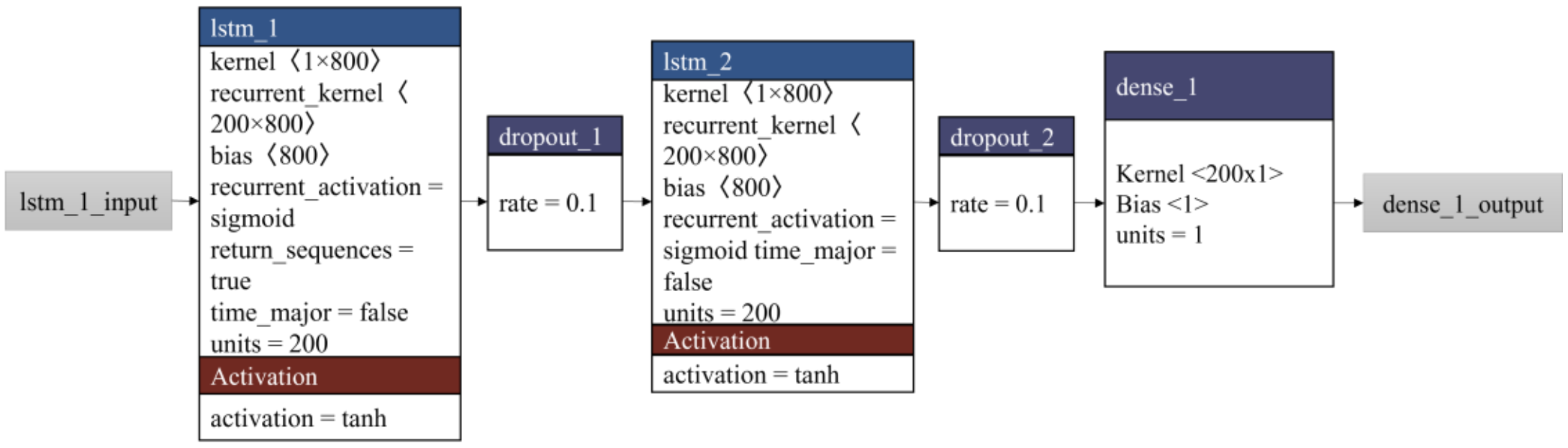

**Figure 3.** Architecture of the proposed forecasting model, REDf

## Model Evaluation Metrics

This section describes the process and metrics of evaluating the performance of the proposed LSTM model for predicting energy demand in a smart power grid. The model was trained using historical energy demand data and evaluated using three performance metrics: mean absolute error (MAE), coefficient of determination ($R^2$), and root mean squared error (RMSE).

The MAE is a measure of the difference between the predicted energy demand and the actual energy demand. It is calculated as the average absolute difference between the predicted and actual values. Mathematically, it is defined as equation (2):

$$MAE = \frac{1}{n}\sum_{i=1}^{n}|y_i \;\; \hat{y}_i| \qquad (2)$$

where $n$ is the number of test samples, $y_i$ is the actual energy demand, and $\hat{y}_i$ is the predicted energy demand. The smaller the MAE, the better the model's performance.

In a linear regression model, the coefficient of determination ($R^2$) represents the proportion of the variance in the dependent variable explained by the independent variables. R-squared values range from 0 to 1, with greater values indicating a superior model fit to the data. A higher R-squared value indicates a better fit and a stronger relationship between the independent and dependent variables when used to evaluate the current model. Mathematically, it is defined as equation (3):

$$R^2 = 1 - \frac{\sum_{i=1}^{n}(y_i - \hat{y}_i)^2}{\sum_{i=1}^{n}(y_i - \bar{y})^2} \tag{3}$$

where $y_i$ is the actual value of the dependent variable, $\hat{y}_i$ is the predicted value of the dependent variable, and $\bar{y}$ is the mean of the dependent variable.

Another evaluation metric used in this study is root mean squared error (RMSE). It measures the average magnitude of the error in the predictions of a model. RMSE calculates the difference between a dataset's actual and predicted values and then takes the square root of the average of those differences. A lower RMSE value indicates that the model better fits the actual values and has higher accuracy in making predictions. Mathematically, it is defined as equation (4):

$$\text{RMSE} = \sqrt{\frac{1}{n}\sum_{i=1}^{n}(y_i - \hat{y}_i)^2} \tag{4}$$

Here, $y_i$ is the actual value, $\hat{y}_i$ is the predicted value, and $n$ is the number of observations.

## Model Deployment

Deploying the proposed LSTM model for securely predicting energy demand in a smart power grid requires appropriate tools and techniques. This section describes the technique of deploying the model securely.

First, TensorFlow (1) is used to serve the model. TensorFlow provides several security features that can be used to protect the model during deployment. For example, TensorFlow can encrypt model data and communicate between the model and other systems. TensorFlow also allows the ability to authenticate users and devices accessing the model. Then, Apache Kafka (18) is used to deploy the model securely. Apache Kafka is a message queuing system that handles high-throughput data streams. It is used to send real-time energy demand data to the model and also receives predictions from it. Apache Kafka provides built-in security features such as encryption and authentication, which protect the data during transmission. To secure the communication between the model and other systems, SSL (14) is used in the proposed model. SSL ensures that all data transmitted between the systems is encrypted and that only authorized systems can access the model. Fig. 4 shows the proposed system's architecture of the model deployment phase. The figure shows that Kafka handles the application, user query, communication between the model server, and prediction output. Users place the prediction requests in the Kafka application, and then the Kafka application forwards the requests to the Tensorflow model server. Then the model server returns the predicted result and handles the response to the Kafka application, which is served to the users. All the communication between the Kafka application and the Tensorflow model server is encrypted with the SSL service. The prototype front end of the application is shown in Fig. 5. This figure shows the interface for taking input and giving prediction output from the Tensorflow model server.

In a nutshell, deploying the proposed LSTM model for securely predicting energy demand in a smart power grid requires using appropriate tools and techniques such as TensorFlow and Apache Kafka. TensorFlow provides built-in security features such as encryption and authentication while serving the model. Apache Kafka also provides built-in security features such as encryption and authentication for data transmission on the application side. Secure communication protocols, such as SSL, are also used to encrypt the data transmitted between systems.

## Experimental Setup

We implemented our proposed approach using Python version 3.6.5 as the primary programming language. The experiments were conducted on a computer with a Ryzen 7 processor, 24GB RAM, and a GTX 1650 GPU. The computer was running Windows 10 as the operating system. We used Jupyter as our integrated development environment (IDE) to develop and test the model.

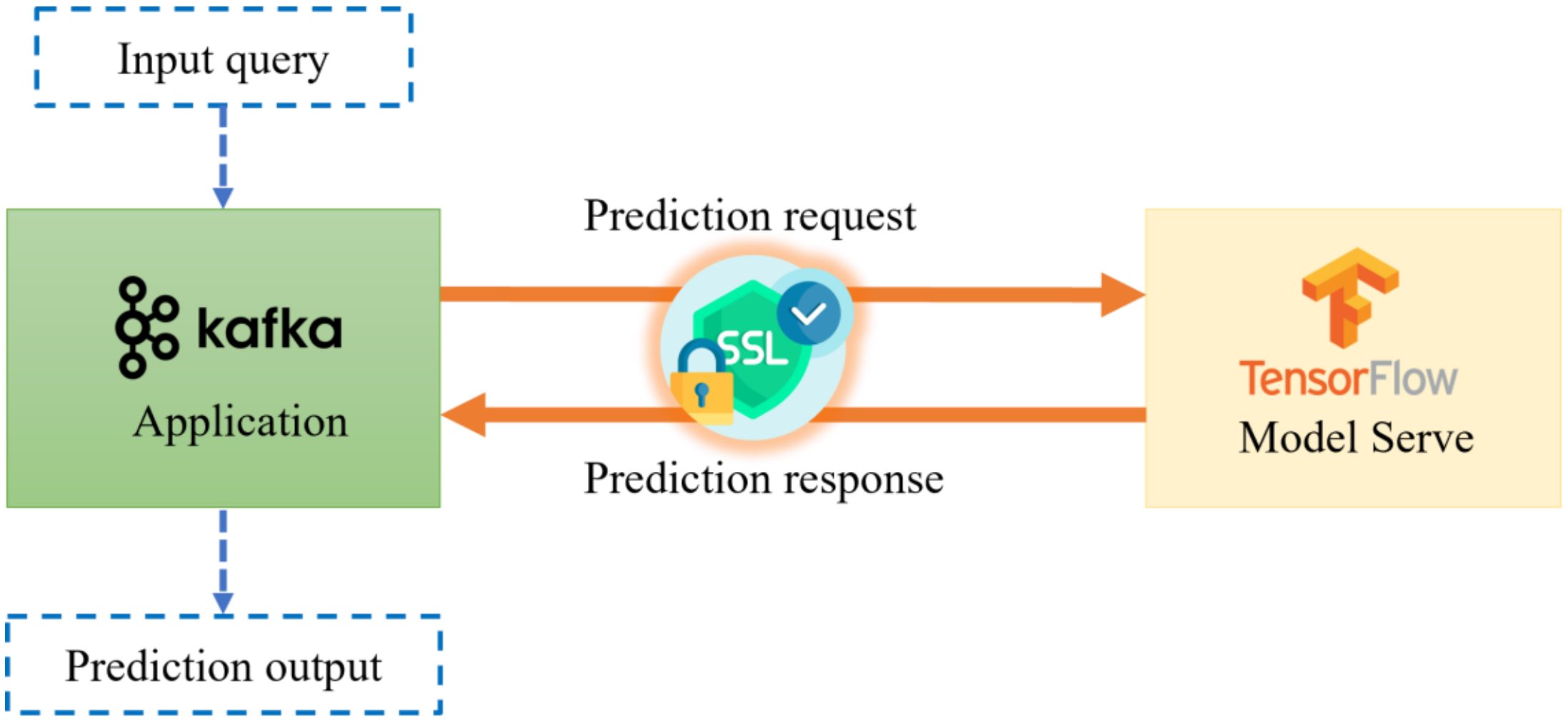

**Figure 4.** Model deployment architecture of the proposed system

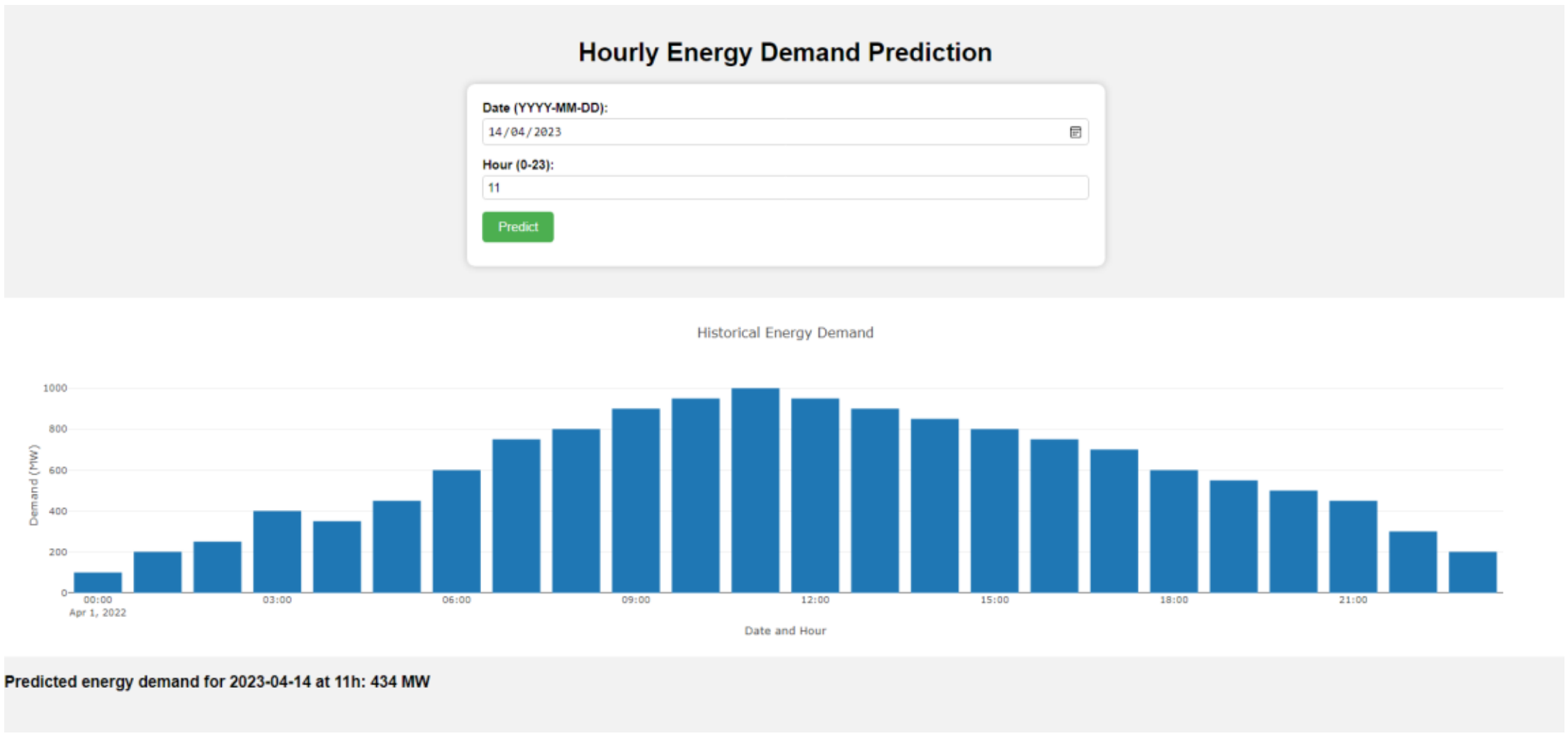

**Figure 5.** Interface of the model deployment application

To implement and train our model, we utilized TensorFlow and Keras, two popular deep-learning frameworks widely used in the research community. TensorFlow provided us with the tools to develop our deep learning model, and Keras allowed us to easily create, compile, and train the model.

Apache Kafka, an open-source distributed event streaming platform, facilitates data streaming. This helped us handle high volumes of data and ensure efficient data processing. For model deployment purposes, we used ModelServe, a high-performance model serving solution that allowed us to quickly and easily deploy our trained model to production environments.

## RESULTS AND DISCUSSION

The proposed deep learning-based approach for predicting energy demand was evaluated using four datasets of historical energy demand data from different energy companies. The datasets consisted of hourly energy demand data for a certain period. The data was divided into a training set, which consisted of 80% of the data, and a test set, which consisted of 20% of the data for all the datasets. The datasets were also trained and tested with three other state-of-the-art machine learning models, namely support vector regression (SVR) (29; 25), random forest regression (RFR) (Khan et al.; 45), and Facebook Prophet (50). The performances of these models are also compared with that of the proposed REDf model.

## Experimental Result Analysis

The deep learning model, REDf was trained using the long short-term memory (LSTM) network architecture with 200 units. The model was trained using the Adam optimization algorithm with a learning rate 0.001. The training process took approximately 25 minutes on the experimental machine.

The performance of the model was evaluated using three metrics: mean absolute error (MAE), root mean squared error (RMSE), and coefficient of determination ($R^2$). The MAE measures how close the predicted values are to the true values, the $R^2$ measures how well the model fits the data, and the RMSE measures the average magnitude of the error in the predictions of a model. The experimental results are shown in Table 3.

**Table 3.** Experimental results for all the evaluation metrics of all models and dataset

| Model-Dataset | $R^2$ | MAE | RMSE |
|---|---|---|---|
| REDf-AEP | 0.983 | 0.015 | 0.024 |
| REDf-COMED | 0.979 | 0.014 | 0.022 |
| REDf-DAYTON | 0.980 | 0.015 | 0.023 |
| REDf-PJME | 0.985 | 0.014 | 0.020 |
| SVR-AEP | 0.982 | 159.269 | 346.603 |
| SVR-COMED | 0.958 | 149.045 | 471.277 |
| SVR-DAYTON | 0.976 | 11.064 | 24.873 |
| SVR-PJME | 0.726 | 1878.685 | 3382.786 |
| Prophet-AEP | 0.052 | 2018.417 | 2522.092 |
| Prophet-COMED | -0.021 | 1782.898 | 2321.032 |
| Prophet-DAYTON | -9.939 | 0.602 | 0.632 |
| Prophet-PJME | -3.298 | 0.359 | 0.407 |
| RFR-AEP | 0.133 | 1926.917 | 2412.619 |
| RFR-COMED | 0.170 | 1613.671 | 2099.070 |
| RFR-DAYTON | 0.065 | 300.336 | 380.377 |
| RFR-PJME | 0.047 | 4890.830 | 6309.890 |

Based on the results presented in Table 3, it is evident that the proposed REDf model achieved high accuracy in the predictions, with MAE ranging from 1.4% to 1.5% across all evaluated datasets, $R^2$ ranging from 97.9% to 98.5% for different datasets, and RMSE of approximately 0.02 across all evaluation datasets. In contrast, the state-of-the-art Support Vector Regression (SVR) model exhibited a good fit for the data, but its performance was inconsistent across all datasets. Despite achieving a good fit for the data, the SVR model's predicted values were not close to the actual data, as evidenced by the high MAE scores. Facebook's Prophet, another state-of-the-art forecasting model, also demonstrated poor performance for all datasets in all evaluation metrics. This model achieved a positive $R^2$ value only for the AEP dataset, while obtaining negative values for other datasets, indicating poor fit and poor predictive capability. The model also exhibited very high scores for MAE and RMSE, signifying poor accuracy in predicting outcomes and significant average error. A similar trend in the Random Forest Regression (RFR) model is observed as the $R^2$ score ranges from 4% to 17% among different datasets. In the case of the RFR model, the high MAE and RMSE values suggest that the model struggled to capture the underlying patterns and trends in the data accurately. This could be attributed to several factors. Firstly, the RFR model may not have been able to effectively handle the temporal dependencies and dynamics present in the time series data. Unlike the REDf model, specifically designed for time series forecasting, the RFR model might not have been able to exploit the sequential nature of the data. Secondly, the RFR model might have been affected by outliers or noise in the data. Random Forest models are generally robust to outliers, but extremely large or influential outliers can still impact the model's performance, leading to higher error values. The experimental results can be further corroborated with visual representations of the models' actual versus predicted data plots. Fig. 6, Fig. 7, Fig. 8, and Fig. 9 show the actual versus predicted data plots for AEP, COMED, DAYTON, and PJME datasets respectively. The high-resolution versions of these figures are available at the following link, https://ping543f.github.io/ren_energy/

The plots depict hourly energy demand over time, with the x-axis representing the time frame and the y-axis representing energy demand. The green line depicts the actual energy demand, while the red line shows the predicted energy demand by the proposed REDf model. For the SVR and RFR models, the red dots indicate actual data points, and the blue dots represent predicted data points. For the Facebook Prophet model, the black dots represent actual data, and the blue line shows the predicted data by the model. Analysis of the plots indicates that the difference between the actual and predicted energy demand is minimal for the proposed REDf model across all datasets. Conversely, the difference between the actual and predicted energy demand is significantly high for the SVR, RFR, and Prophet models in almost all datasets, except for the SVR model in the Dayton dataset. The actual versus predicted plots for the proposed REDf model demonstrate that the predicted energy demand values are very close to the actual values, indicating a good fit for the data and accurate forecasting of energy demand data. Furthermore, the proposed model showed no sign of overfitting and underfitting at the end of the training epochs. In this study, the performance of the developed model was carefully evaluated to assess the presence of overfitting and underfitting phenomena. Figure 10 shows the loss curves for training and testing phases of the model for different datasets employed in this study. Overfitting occurs when a model performs exceptionally well on the training data but fails to generalize to new, unseen data. On the other hand, underfitting happens when a model lacks the complexity to capture the underlying patterns in the data and performs poorly on both the training and test sets.

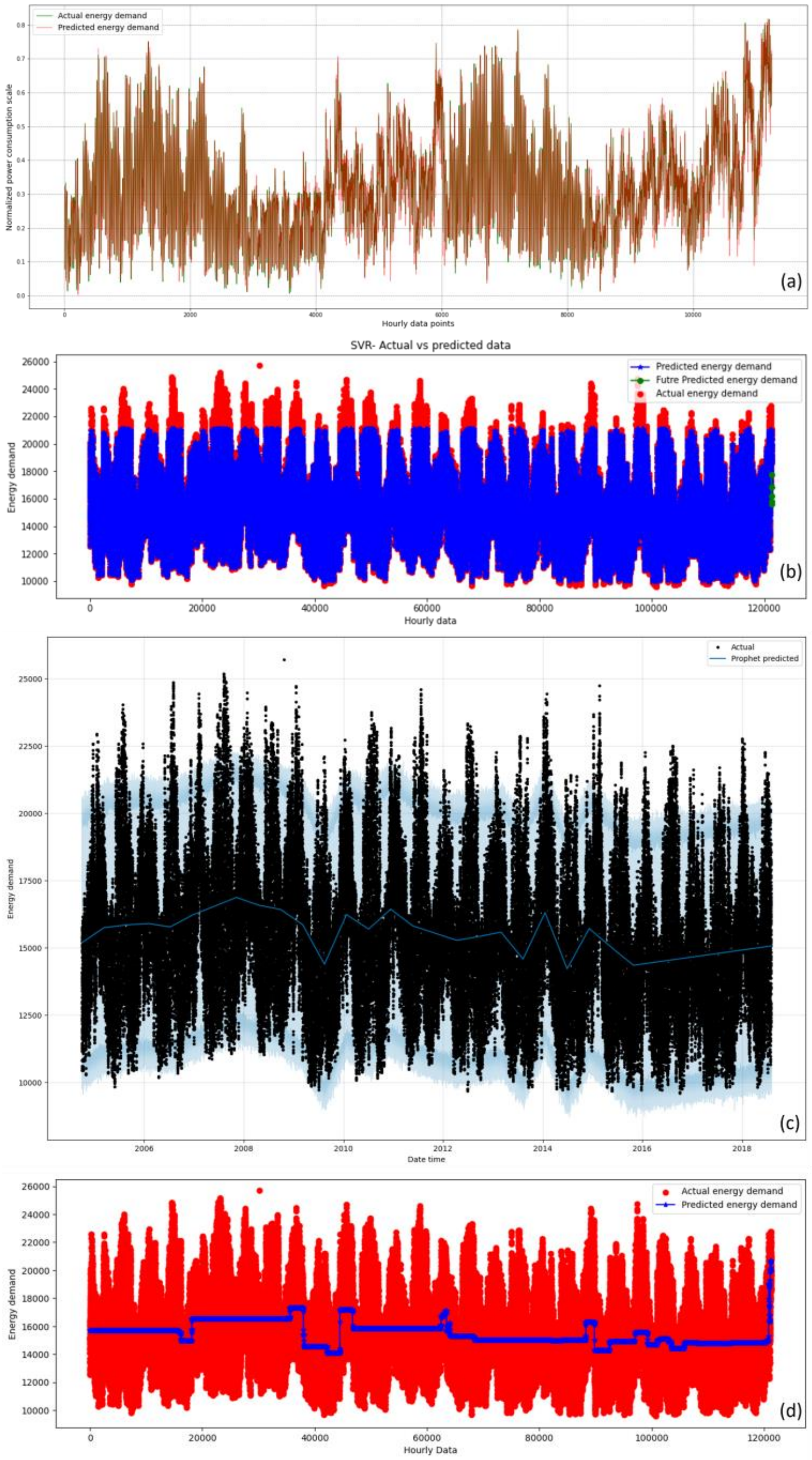

**Figure 6.** Actual vs. Predicted data for different models for AEP dataset. (a) Proposed REDf model, (b) SVR Model, (c) Facebook Prophet model, and (d) RFR model

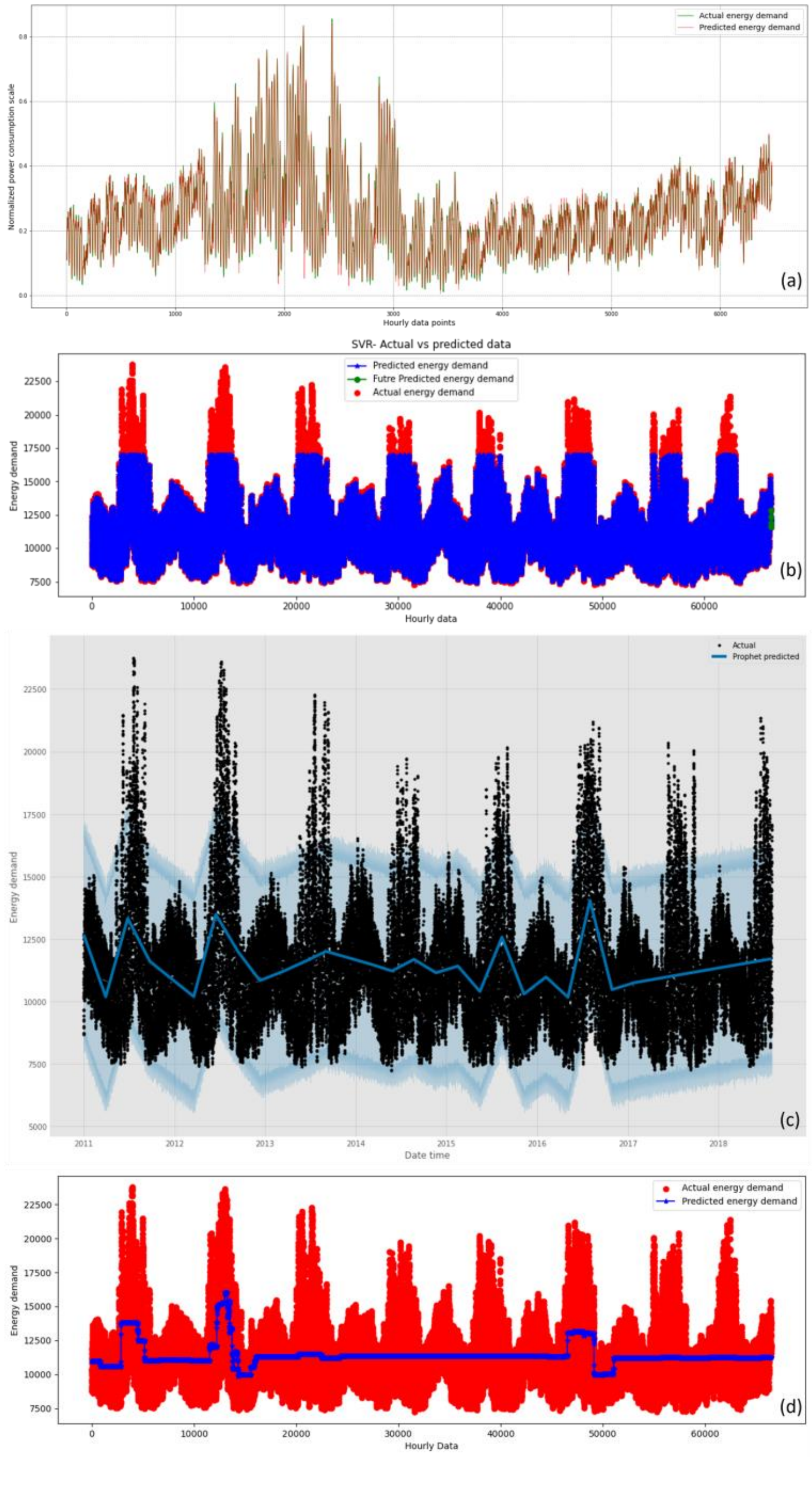

**Figure 7.** Actual vs. Predicted data for different models for COMED dataset. (a) Proposed REDf model, (b) SVR Model, (c) Facebook Prophet model, and (d) RFR model

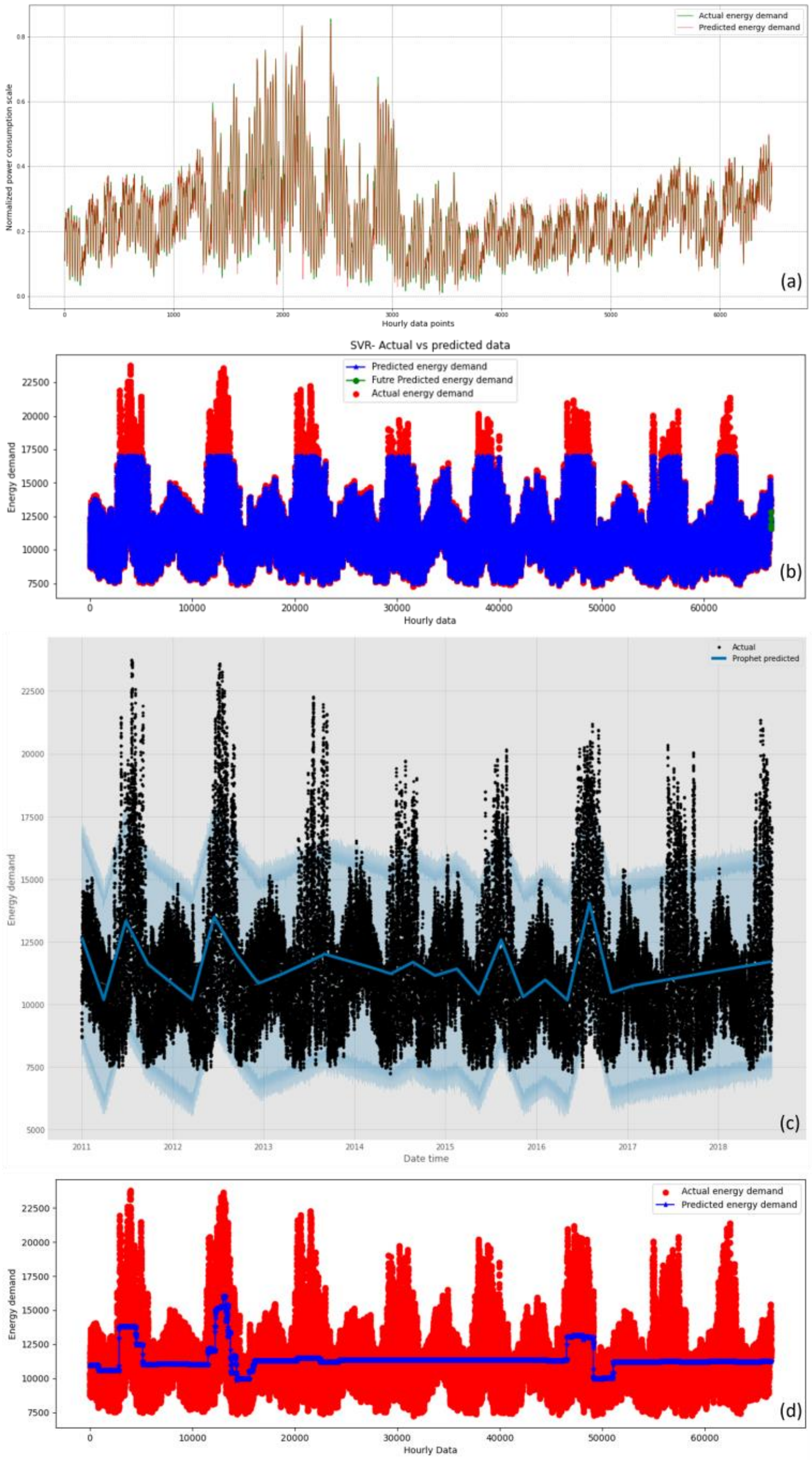

**Figure 8.** Actual vs. Predicted data for different models for DAYTON dataset. (a) Proposed REDf model, (b) SVR Model, (c) Facebook Prophet model, and (d) RFR model

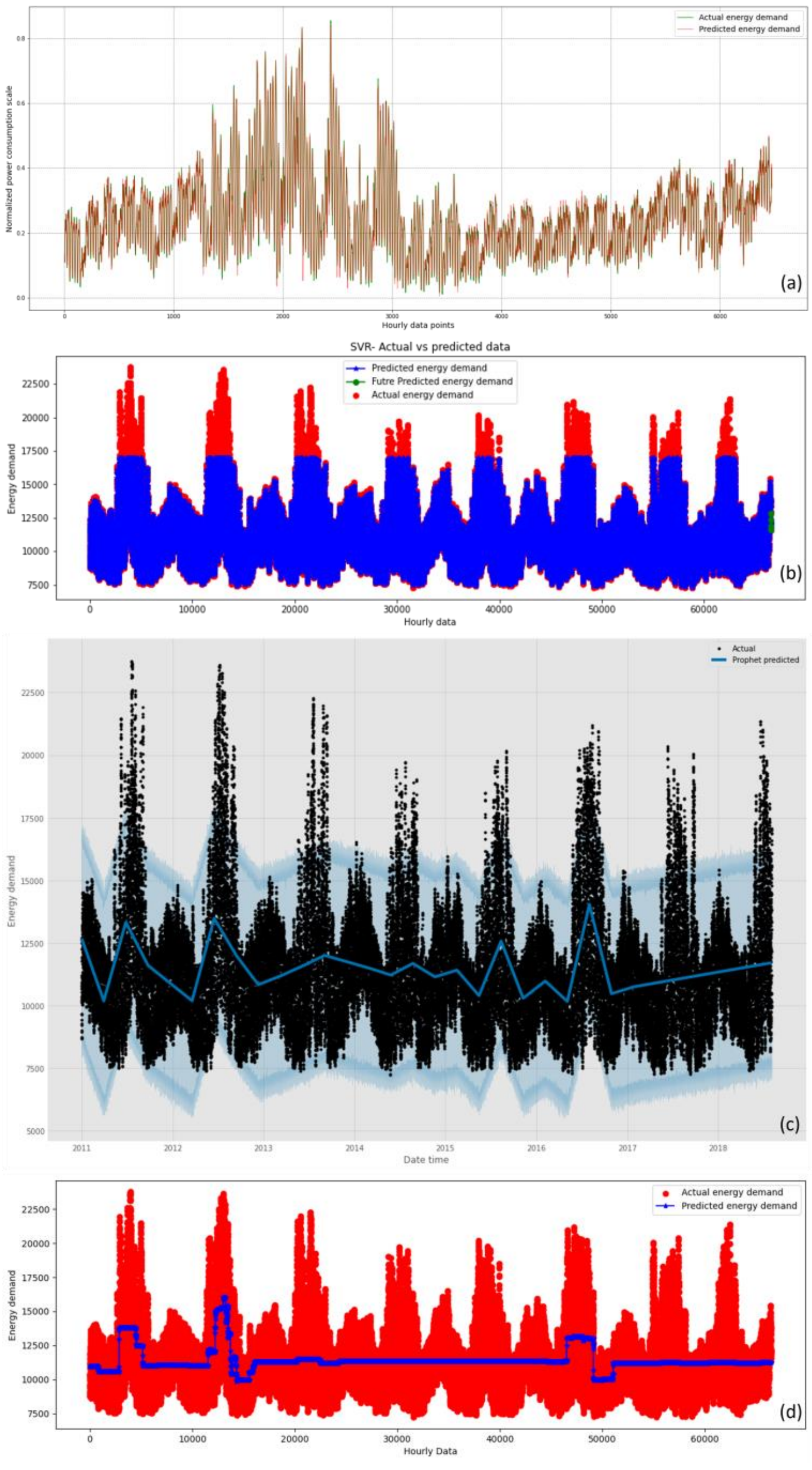

**Figure 9.** Actual vs. Predicted data for different models for PJME dataset. (a) Proposed REDf model, (b) SVR Model, (c) Facebook Prophet model, and (d) RFR model

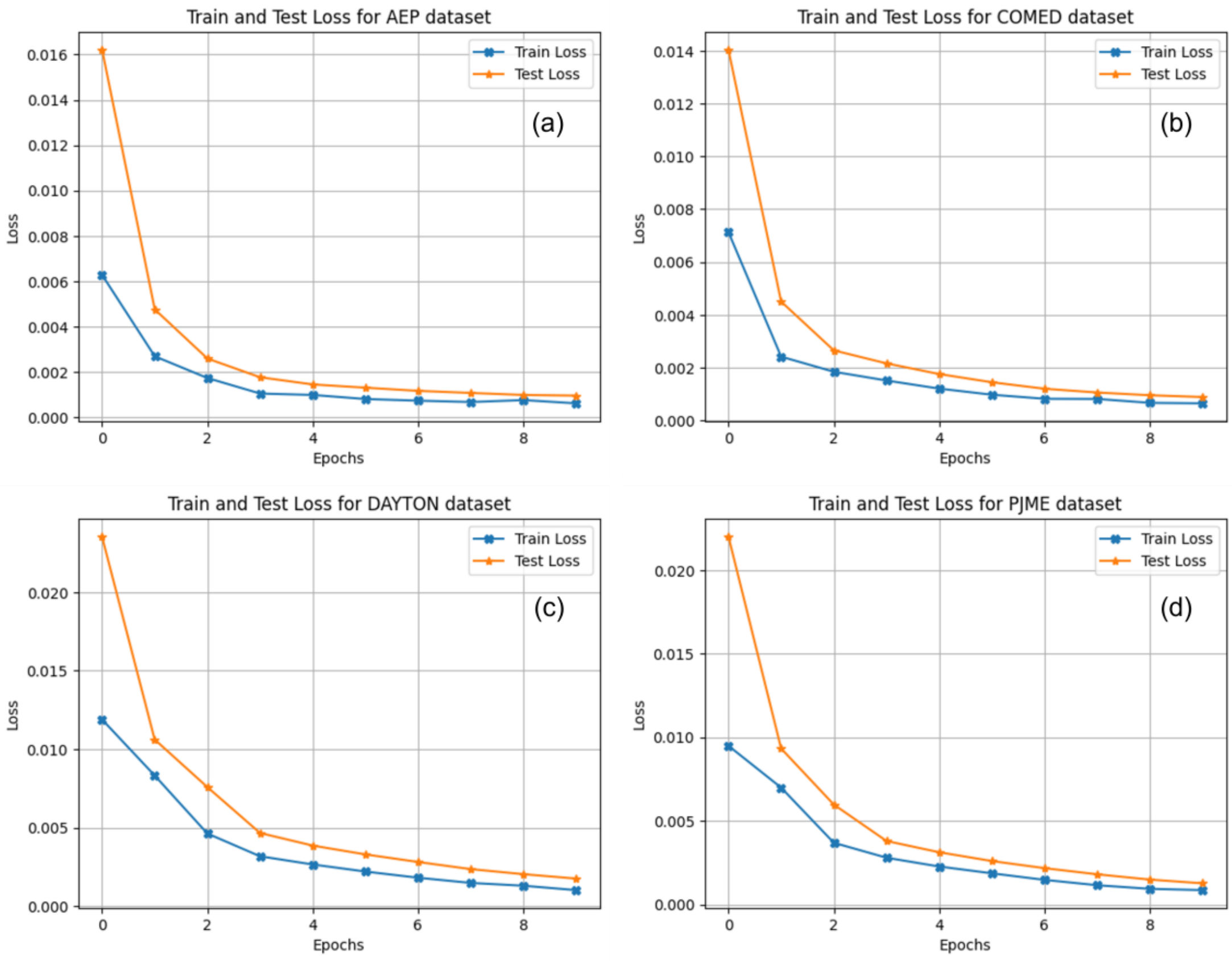

**Figure 10.** Loss curves of training and testing phases for the proposed REDf model in different datasets. (a) represents the curves for AEP dataset, (b) represents COMED dataset, (c) represents DAYTON dataset and (d) represents PJME dataset.

The model demonstrated robust performance in our experiments without showing signs of overfitting or underfitting. This was evident from the consistent and comparable performance metrics achieved on the training and test datasets. The absence of overfitting can be attributed to including dropout layers in the model, which regularizes the training process by randomly dropping a fraction of the units, preventing the model from relying too heavily on specific features. Additionally, by selecting an appropriate model architecture and hyperparameters through a systematic grid search technique, we ensured that the model had the necessary complexity to capture the underlying patterns in the data without excessive complexity that could lead to overfitting.

The absence of underfitting indicates that the model could adequately capture the relevant information from the training data, allowing it to generalize well to unseen data. This suggests that the chosen model architecture and hyperparameters were suitable for the given task and dataset, striking a balance between simplicity and complexity.

Based on the experimental results, the proposed REDf model is highly accurate in forecasting energy demand. The model's predicted values align with the original energy demand values. This indicates that the model is a good fit for the data and can be relied upon for accurate predictions of energy demand. Demand forecasting accuracy plays a significant role in integrating renewable energy sources and achieving sustainable development goals (SDGs) in the smart grid. Accurate demand prediction allows grid operators to anticipate load variations, allocate resources effectively for frequency regulation services, and optimize resource allocation by aligning them with expected load patterns. This improves the scheduling and dispatch of frequency regulation resources, minimizes the need for corrective actions, and reduces the risk of frequency deviations. Moreover, accurate load forecasting enhances market participation, enabling market participants to make informed bidding decisions based on expected load fluctuations, resulting in more competitive and accurate bids. Ultimately, load forecasting accuracy contributes to overall grid stability and reliability by allowing proactive measures to balance supply and demand, thereby reducing the likelihood of frequency deviations and ensuring the reliable operation of power systems.

## Comparative Analysis

It is important to note that the results are based on a specific dataset and architecture, and the performance of the proposed approach may vary depending on the type of data and the specific implementation details. However, the results demonstrate the potential of the proposed approach for predicting short-term energy demand in a smart power grid. The results achieved from the experiment can be compared to other recent works in predicting energy demand in smart power grids.

One of the most closely related studies is by Amalou et al. (8), who used a similar approach to deep learning with Long Short-Term Memory (LSTM) networks to predict energy demand. In omparison to this study, the proposed REDf model achieved better performance in terms of MAE, $R^2$, and RMSE. Specifically, the proposed model achieved a mean absolute error of 1.4%, which is significantly lower than the MAE of 0.021 reported in this study. Additionally, the proposed model achieved a higher $R^2$ value, indicating a better fit to the data, and a lower RMSE value, indicating better accuracy in predicting energy demand.

Similarly, compared to the study by Alrasheedi et al. (7), the proposed REDf model achieved better performance in terms of RMSE and $R^2$. Alrasheedi et al. achieved an RMSE of 0.8168 and a $R^2$ of 0.973 on their test set, while the proposed REDf model achieved a lower RMSE and a higher $R^2$ value . This indicates that the proposed model is more accurate in predicting energy demand and better fits the data.

A recent study by Shachee et al. (47) proposed a hybrid deep learning-based model of LSTM and RNN that utilizes historical load data to predict energy demand. They achieved an RMSE of 0.6 on their test set. Our model achieved better performance in this evaluation metric compared to this study.

Taleb et al. (49) presented a hybrid model that combines standard neural networks with an automatic weight update process, achieving a mean absolute error (MAE) of 372.08 in energy demand prediction. Mubashar et al. (38) proposed a method for load forecasting using LSTM models and compared its performance with two commonly used techniques, ARIMA and exponential Smoothing. Their proposed method outperformed the other two, achieving an MAE of 2.44736176. Rosato et al. (46) presented a novel deep learning approach using Convolutional Neural Network and Long Short-Term Memory models, achieving the lowest RMSE of 2.252 for the baseline 1-day forecast. Nguyen et al. (39) proposed an electricity demand forecasting method based on the LSTM deep learning model, achieving an RMSE of 9.63. Pramono et al. (44) proposed a method for short-term load forecasting using a wavenet-based model that employs dilated causal residual CNN and LSTM layers, achieving RMSE and MAE equal to 203.23 and 142.23 for ENTSO-E dataset 1 and 292.07 and 196.95 for ENTSO-E dataset 2. The proposed methods have demonstrated their potential for supporting energy management and demand response programs in hybrid energy systems.

The performance of the proposed model was compared with other recent studies in predicting energy demand in smart power grids. The proposed model outperforms the other models regarding MAE, $R^2$, and RMSE in all the evaluated datasets. The LSTM-RNN (47) model did not report the MAE and $R^2$ values mentioning only RMSE. The CNN-GRU (7) model achieved a high $R^2$ score but did not report the MAE and RMSE values. On the other hand, the LSTM and GRU models reported MAE and $R^2$ values, but their performance was inferior to the proposed model. Apart from these, most studies did not report $R^2$ values.

Various deep learning-based methods have been proposed to accurately forecast energy demand, including standard neural networks with an automatic weight update process, LSTM models, CNN and LSTM models, and wavenet-based models. These models have been shown to outperform traditional methods such as ARIMA and exponential smoothing, achieving lower MAE and RMSE values. However, our proposed model significantly outperforms all the existing models, achieving an exceptionally low MAE of 0.015 and RMSE of 0.02, demonstrating its potential to revolutionize the energy sector by providing more accurate energy demand forecasting.

The proposed model for predicting the demand for energy in a smart grid can also assist in realizing substantial environmental advantages while advancing several sustainable development goals (SDGs). First and foremost, this paradigm can help fulfill SDG 7: Access to Affordable and Clean Energy. Utility companies can better manage their renewable energy resources and lessen their dependency on fossil fuels by precisely anticipating the demand for renewable energy. This will result in a more ecologically friendly and sustainable energy system. This may facilitate the transition to a low-carbon economy, improve air quality, and cut greenhouse gas emissions. As a result, people's access to and affordability of energy, particularly in low-income areas, may improve.

In addition, this demand model for renewable energy might help with SDG 9: Industry, Innovation, and Infrastructure. Utility companies and other stakeholders can build sustainable infrastructure and encourage innovation in the energy industry by offering accurate and trustworthy estimates of the demand for renewable energy. New technologies and business models may be created as a result, which might hasten the uptake of renewable energy sources and encourage their use in a sustainable and efficient manner.

Thirdly, this model can help achieve SDG 13: Climate Action. Predictive models for renewable energy demand can aid in the fight against climate change and its effects by encouraging renewable energy sources and lowering dependency on fossil fuels. To mitigate the effects of climate change, such as more frequent and severe weather events; this can involve lowering greenhouse gas emissions, enhancing air quality, and enhancing air quality.

Achieving various sustainable development objectives relating to access to affordable and clean energy, innovation and infrastructure, and climate action can be facilitated by developing precise forecast models for renewable energy demand in smart grids. Our proposed model, which uses deep learning, LSTM networks, and data pre-processing approaches, performed better than recent research in this sector. This shows that our method can be an efficient way to estimate energy demand in smart power grids and might have significant economic and environmental benefits by encouraging the adoption of renewable energy sources and lowering dependency on fossil fuels. As a result, our work contributes significantly to the ongoing efforts to create sustainable energy systems that can support a more equitable future and less harmful to the environment.

## Environmental Benefits

Using machine learning methods like deep learning to predict energy consumption has the potential to improve smart power grids significantly. Power grid administrators may maximize the distribution and use of renewable energy sources, minimizing reliance on non-renewable sources and encouraging the integration of clean energy by accurately projecting energy demand. As a result, customers and the environment could benefit from decreased prices, increased efficiency, and better control over power networks. The generalization and prediction abilities of the suggested method have been shown to be strong. This strategy can support the international effort to combat climate change and achieve sustainable development by incorporating the principles of SDGs 7 (affordable and clean energy), 9 (industry, innovation, and infrastructure), and 13 (climate action).

## Challenges

Although the proposed method for predicting energy demand using deep learning has the potential to improve the integration of renewable energy sources and optimize the efficiency of power infrastructures, it is not without obstacles. The data availability and quality required for training deep learning models is a significant challenge. It is possible that historical energy demand data are unavailable or insufficient, which can compromise the accuracy of the model's predictions. Another obstacle is the high computational requirements and lengthy nature of deep learning model training. This can be problematic when working with enormous datasets or multiple variables. In addition, the interpretability of the model can be problematic, as the inner workings of deep learning models can be challenging to comprehend, limiting their transparency and accountability. It is crucial to successfully address these obstacles to implement the proposed approach in smart power infrastructures.

The results of this work can be summed up as follows: (1) The suggested model outperformed other recent efforts in anticipating energy consumption in smart power grids in terms of performance. (2) Deep learning with LSTM networks and data pre-processing methods successfully anticipated energy consumption in smart power grids. (3) The suggested model has the ability to help achieve environmental benefits and sustainable development goals, as evidenced by its accuracy in anticipating energy consumption in smart power networks. (4) The suggested model can aid in the more effective use of renewable energy sources by better forecasting energy demand and lowering the requirement for environmentally hazardous non-renewable energy sources. (5) A more sustainable use of natural resources can be achieved by reducing energy waste and using energy more efficiently.

## CONCLUSION

This study proposes an LSTM-based deep learning model for forecasting energy demand in smart power grids. Four distinct datasets, including AEP, COMED, DAYTON, and PJME, were used to evaluate the proposed model, which employed data pre-processing techniques. The model obtained excellent results across all datasets, with a mean absolute error (MAE) of between 1.4% and 1.5%. Moreover, the model attained the highest $R^2$ score of all datasets evaluated, 98.5%. These results indicate that the proposed model accurately predicts energy demand in smart power grids. The development of accurate energy demand prediction models is essential for attaining multiple sustainable development objectives relating to affordable and clean energy, innovation and infrastructure, and climate action. Predictive models for energy demand can help mitigate the adverse effects of climate change and promote a more sustainable and environmentally benign energy system by reducing reliance on fossil fuels and promoting the use of renewable energy sources. Our proposed model significantly contributes to ongoing efforts to develop sustainable energy systems that can support a more equitable and environmentally favourable future.

The proposed model performed exceptionally well in anticipating the short-term energy demand in smart power grids. Combining deep learning with LSTM networks and data pre-processing techniques proved to be an effective method for accurately predicting the demand for renewable energy. These results demonstrate the practical applicability of the proposed model, providing utilities and other stakeholders with a valuable tool to manage their renewable energy resources and promote more efficient and sustainable energy use. Increasing the accuracy of predictions and incorporating other factors, such as weather conditions and renewable energy generation forecasts, can be the subject of additional study. In addition, it would be intriguing to investigate how this technique could be combined with other technologies, such as IoT and blockchain, to create a more robust and effective smart grid system.

## AUTHOR CONTRIBUTION

Conceptualization, M.S.U.M., J.S. and M.I.I.; Investigation, M.S.U.M. and M.M.; Methodology, M.S.U.M; Data Curation, M.S.U.M.; Writing—Original Draft Preparation, M.S.U.M. and M.I.I; Supervision, J.S.; Funding Acquisition, M.S.H.L. and R.S.R; Project Administration, M.S.U.M. and J.S.; Writing—Review and Editing, M.S.H.L. and R.S.R. All authors have read and agreed to the published version of the manuscript.

## DATA AVAILABILITY STATEMENT

The data and codes are available at the following Github repository, https://github.com/ping543f/ren-energy

## ACKNOWLEDGMENTS

This research was supported by the Post Graduate Research Scheme of Universiti Malaysia Pahang entitled: An automated knowledge-based framework for EDLC using deep learning approach, PGRS220337.